%% file: main.tex
\documentclass{article}

     \PassOptionsToPackage{numbers, compress}{natbib}

\usepackage[final]{neurips_fitml_2024}

\usepackage[utf8]{inputenc} 
\usepackage[T1]{fontenc}    
\usepackage{hyperref}       
\usepackage{url}            
\usepackage{booktabs}       
\usepackage{multirow}       
\usepackage{amsfonts}       
\usepackage{nicefrac}       
\usepackage{microtype}      
\usepackage{xcolor}         
\usepackage{amssymb}
\usepackage{amsmath}
\usepackage{mathtools}
\usepackage{amsthm}
\usepackage{colortbl}
\usepackage{makecell}
\usepackage[capitalise,nameinlink]{cleveref}
\usepackage{siunitx}
\usepackage{subcaption}

\crefname{section}{Sec.}{Sec.}
\crefname{thm}{Thm.}{Theorem}
\crefname{appendix}{App.}{Appendices}
\crefname{algorithm}{Alg.}{Algorithms}
\crefname{equation}{Eq.}{Eqs.}
\crefname{figure}{Fig.}{Figs.}
\creflabelformat{equation}{#2\textup{#1}#3}

\newcommand{\bbE}{\mathbb{E}}
\newcommand{\loss}{\ell}
\newcommand{\myvec}[1]{\mbox{$\mathbf{#1}$}}
\newcommand{\myvecsym}[1]{\mbox{$\boldsymbol{#1}$}}
\newcommand{\vtheta}{\mbox{$\myvecsym{\theta}$}}
\newcommand{\vparam}{\vtheta}
\newcommand{\vA}{\mbox{$\myvec{A}$}}
\newcommand{\vB}{\mbox{$\myvec{B}$}}
\newcommand{\vI}{\mbox{$\myvec{I}$}}
\newcommand{\vm}{\mbox{$\myvec{m}$}}
\newcommand{\vv}{\mbox{$\myvec{v}$}}
\newcommand{\dkls}[3]{\mathbb{D}_{\text{KL}}^{#1}[#2 \, \|\, #3]}
\newcommand{\gauss}{\mbox{${\cal N}$}}

\title{Variational Low-Rank Adaptation Using IVON}

\author{%
  Bai Cong$^{1,2}$ \quad Nico Daheim$^4$ \quad Yuesong Shen$^{3,5}$ \quad Daniel Cremers$^{3,5}$ \quad Rio Yokota$^1$ \\
  \textbf{Mohammad Emtiyaz Khan}$^2$ \quad \textbf{Thomas Möllenhoff}$^2$ \\
  \\
  $^1$Institute of Science Tokyo \quad $^2$RIKEN Center for AI Project \quad $^3$Technical University of Munich \\
  $^4$Technical University of Darmstadt \quad $^5$Munich Center for Machine Learning
}

\begin{document}

\maketitle

\begin{abstract}

   We show that variational learning can significantly improve the accuracy and calibration of Low-Rank Adaptation (LoRA) without a substantial increase in the cost.~We replace AdamW by the Improved Variational Online Newton (IVON) algorithm to finetune large language models. For Llama-2 with 7 billion parameters, IVON improves the accuracy over AdamW by $2.8\%$ and expected calibration error by $4.6\%$.
   The accuracy is also better than the other Bayesian alternatives, yet the cost is lower and the implementation is easier.
   Our work provides additional evidence for the effectiveness of IVON for large language models.~The code is available at \url{https://github.com/team-approx-bayes/ivon-lora}.
\end{abstract}

\section{Introduction}

Bayesian methods are expected to improve the accuracy and calibration performance of Large Language Models (LLMs) on downstream tasks, but they rarely succeed at such massive scale and, even when they do, often there is a substantial cost to pay.
This is certainly true for finetuning with Low-Rank Adaptation~\citep{hu2021lora}, where many Bayesian variants have recently been proposed but they all require additional computations compared to standard finetuning practices. For example, the SWAG-LoRA method~\citep{onal2024gaussian} needs additional computation to obtain an estimation of the posterior. LoRA ensemble~\citep{wang2023lora} requires multiple LoRA checkpoints to be trained.
Methods such as Laplace-LoRA~\citep{yang2023bayesian} require an additional pass through the data to compute a Hessian or Fisher
approximation.
It is then natural to ask whether it is ever possible to use Bayes to improve LoRA without such overheads and increase in the costs. 

Here, we show that the variational (Bayesian) learning can significantly improve both the accuracy and calibration of LoRA finetuning without a substantial increase in the cost. Our proposal is to simply replace the standard optimizers like AdamW by a variational learning algorithm called the Improved Variational Online Newton (IVON) algorithm~\citep{shen2024variational}. IVON uses a nearly identical implementation as AdamW and the swap requires just a few lines of code change.
The main advantage of IVON is that its scale vector, used for the learning rate adaptation, also yields an estimate of posterior variance for free. The only minor overhead is due to sampling from the posterior but we show that this cost is negligible in practice (approximately $1\%$ of the total training time). We achieve significant improvements in performance when finetuning the Llama-2 model with 7 billion parameters on a range of commonsense reasoning tasks: accuracy increases by $2.8\%$ while expected calibration error (ECE) decreases by $4.6\%$. 
The accuracy is also better than the other Bayesian alternatives, yet the cost is much lower and the implementation is easier. Our work provides additional evidence for the work of \citet{shen2024variational}, showing effectiveness of IVON for large deep networks.

\section{Variational low-rank adaptation using IVON}

We will introduce our approach that we call IVON-LoRA. The idea is simple: we replace the standard AdamW optimizer by IVON which optimizes a variational-Bayesian objective. In other words, we switch the standard objective used by AdamW to a variational one. More formally, let us denote the AdamW objective by $\loss(\vparam)$ where $\vparam$ is the vector containing all the entries of LoRA's low-rank parameters (often denoted by $\vA$ and $\vB$). The variational learning
minimizes a different objective where an expectation of $\loss(\vparam)$ over a distribution $q(\vparam)$ is used (shown on the right), 
\begin{equation}
   \min_{\text{\vparam}}  \,\, \loss(\vparam) \quad \text{ versus } \quad
   \min_{q(\text{\vparam})} \,\, \bbE_{q(\text{\vparam})} \left[ \loss(\vparam) \right] + \frac{1}{\lambda} \, \dkls{}{q(\vparam)}{p(\vparam)}.
   \label{eq:AdamWvsIVON}
\end{equation}
IVON uses a Gaussian $q(\vparam) = \gauss(\vm, \text{diag}(\vv))$. The mean $\vm$ plays a similar role to $\vparam$ obtained by AdamW, while the posterior variance vector $\vv$ is an additional quantity. The prior $p(\vparam) = \gauss(0,v_0\vI)$ is a zero mean isotropic Gaussian with a scalar variance $v_0$. A scalar weighting parameter $\lambda$ is used to take care of the data size $N$. This is because
$\loss(\vparam)$ is often an \emph{average} over the whole dataset. Therefore, when using $\lambda = N$, we target the posterior distribution while with larger values we go towards a ``colder'' posterior \cite{zhang2006varepsilon,DBLP:conf/alt/Grunwald12}.

Despite such differences in the objectives, the implementation of IVON is nearly identical to AdamW which makes the replacement easy and can be done by just a few lines of code change. The key point is that estimation of $\vv$ is automatically done through the scale vector that adapts the learning rate. Therefore, posterior variances are obtained for free.
The only additional step is to sample $\vparam \sim \gauss(\vm,\text{diag}(\vv))$ to evaluate the expectation in \cref{eq:AdamWvsIVON}, but its overhead can be reduced by using one Monte-Carlo sample per iteration. For the details, we refer to the original IVON paper by \citet{shen2024variational}. Overall, IVON is a promising alternative to the existing Bayesian approaches that require additional overheads due to either post-processing or
extra training runs.

\section{Experiments}
\label{sec:experiments}
To evaluate the effectiveness of the proposed method,
we use IVON to finetune a pretrained Llama-2 model with 7 billion parameters~\citep{touvron2023llama} on six datasets with commonsense reasoning multiple-choice or true/false questions.
These six datasets are WinoGrande-Small (WG-S), WinoGrande-Medium (WG-M) \citep{sakaguchi2021winogrande}, ARC-Challenge (ARC-C), ARC-Easy (ARC-E) \citep{clark2018think}, OpenBookQA (OBQA) \citep{mihaylov2018can}, and BoolQ \citep{clark2019boolq}.
We evaluate the performance of the trained LoRA adapters by calculating the accuracy and Expected Calibration Error (ECE) on the test set. We also use test Negative Log-Likelihood (NLL) and Brier score because ECE can be sometimes unreliable \citep{baan-etal-2022-stop}.
As for the baseline methods, we compare the performance of IVON-LoRA adapters with LoRA adapters trained using AdamW.
We also consider other Bayesian alternatives,
including Monte Carlo Dropout (MC Dropout) \citep{GaGh16}, Laplace Approximation (LA) \citep{yang2023bayesian}, Stochastic Weight Averaging (SWA) \citep{izmailov2018averaging,onal2024gaussian}, and SWA-Gaussian~(SWAG)~\citep{maddox2019simple,onal2024gaussian}.

IVON is evaluated in two ways at test time: first, by using the prediction just at the mean $\vm$, and second, by using an averaged prediction over 10 samples from the posterior distribution. The two methods are referred to as `IVON@mean' and `IVON', respectively.
For a fair comparison, we use the same number of samples for MC Dropout and SWAG.

The results are summarized in Table \ref{tab:performance}.
First, we observe that IVON, as an alternative to AdamW, significantly improves the generalization of LoRA finetuning.
When evaluated at the mean, IVON outperforms AdamW finetuning and other Bayesian adaptations of LoRA on all datasets in terms of accuracy, often by a large margin.
We also observe that IVON exhibits improved calibration compared to AdamW and MC Dropout, as indicated by the lower ECE, NLL and Brier values.

\input{acc-calib.tex}

Next, we observe that ensembling with samples from IVON's posterior distribution further improves calibration.
When evaluate at an ensemble of 10 samples drawn from the posterior distribution,
IVON outperforms all other methods and is comparable to the best-performing LA (with a Kronecker-factored Hessian) and SWAG on ECE, NLL and Brier.
Notably, IVON achieves this despite using a diagonal Hessian and without an additional pass through the data for computing Hessians at the converged point as in Laplace methods.
With this improvement in calibration, IVON still maintains comparable or better accuracy over other methods.

\begin{figure}[h]
  \centering
  \includegraphics[width=\textwidth]{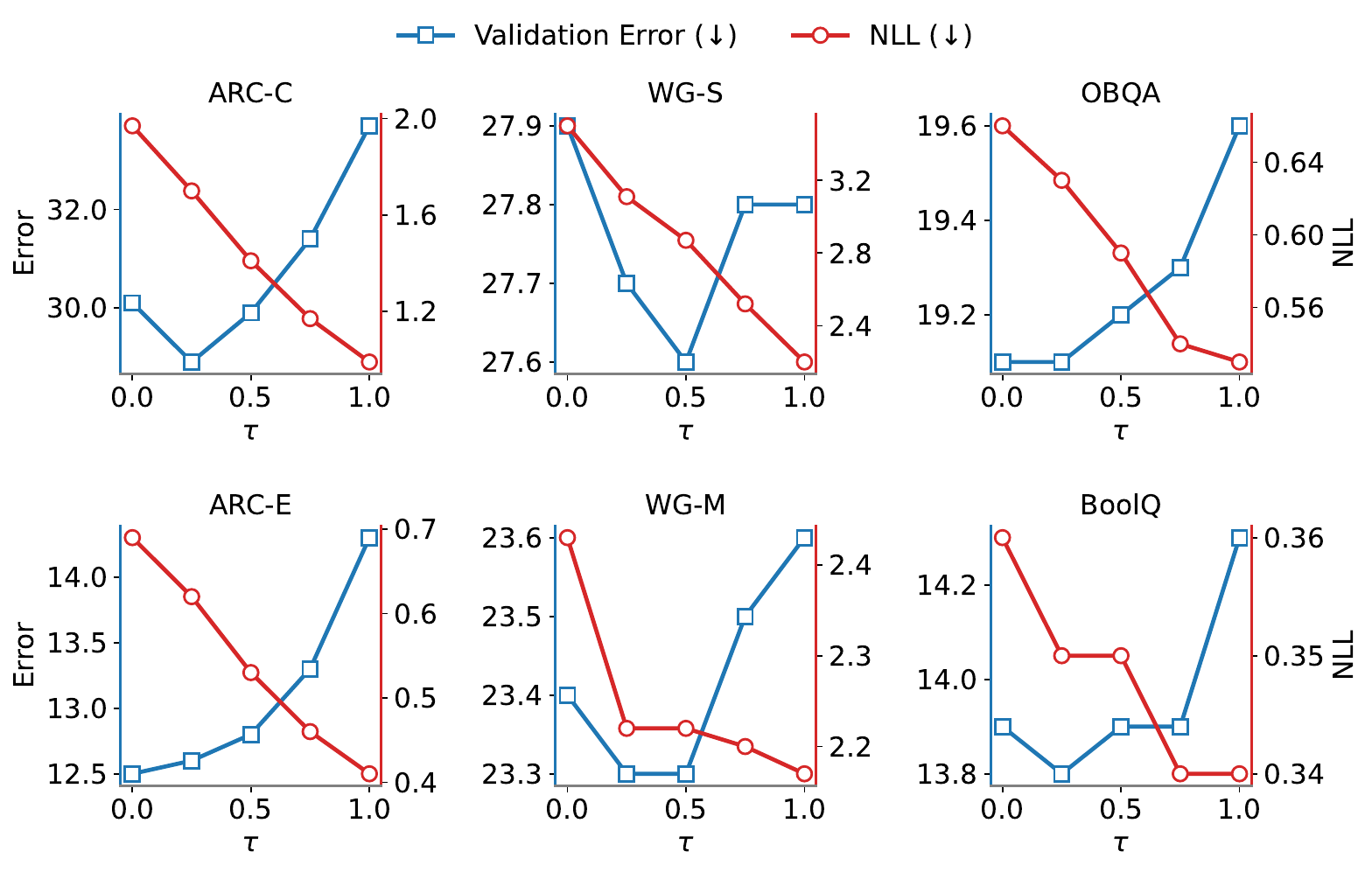}
   \caption{Interpolation between `IVON@mean' and `IVON' enables us to trade-off accuracy for better calibration at test time. Essentially, we use $\gauss(\vparam \,|\, \vm, \text{diag}(\tau\vv))$ with a scalar $\tau \in [0,1]$.
   For $\tau=0$, we recover IVON@mean (leftmost marker) and, for $\tau = 1$, we recover IVON (rightmost marker). Generally, as $\tau$ is increased, the error increases while the NLL decreases. The trend is consistent across datasets (with a few minor exceptions). Metrics are averaged over 3 runs.}
  \label{fig:ablation}
\end{figure}

It is also possible to interpolate between IVON@mean and IVON to achieve the best of both.
Specifically, at test time, we can scale $\vv$ by a scalar $\tau>0$, that is, we predict using parameters sampled from $\mathcal{N}(\vparam \,| \, \vm, \text{diag}(\tau\vv))$. For $\tau=0$, we get IVON@mean and, for $\tau =1$, we get IVON. Increasing $\tau$ then allows us to gradually explore the neighboring solutions around the mean and take advantage of the diversity to improve calibration with a graceful loss of the accuracy.
This is shown in~\Cref{fig:ablation} where we see that, as $\tau$ increases, the error
increases (accuracy decreases) but the NLL decreases (calibration improves). In practice, this simple scaling technique is useful to get a desirable trade-off between accuracy and calibration for specific applications.

\begin{figure}[h]
  \centering
  \includegraphics[width=\textwidth]{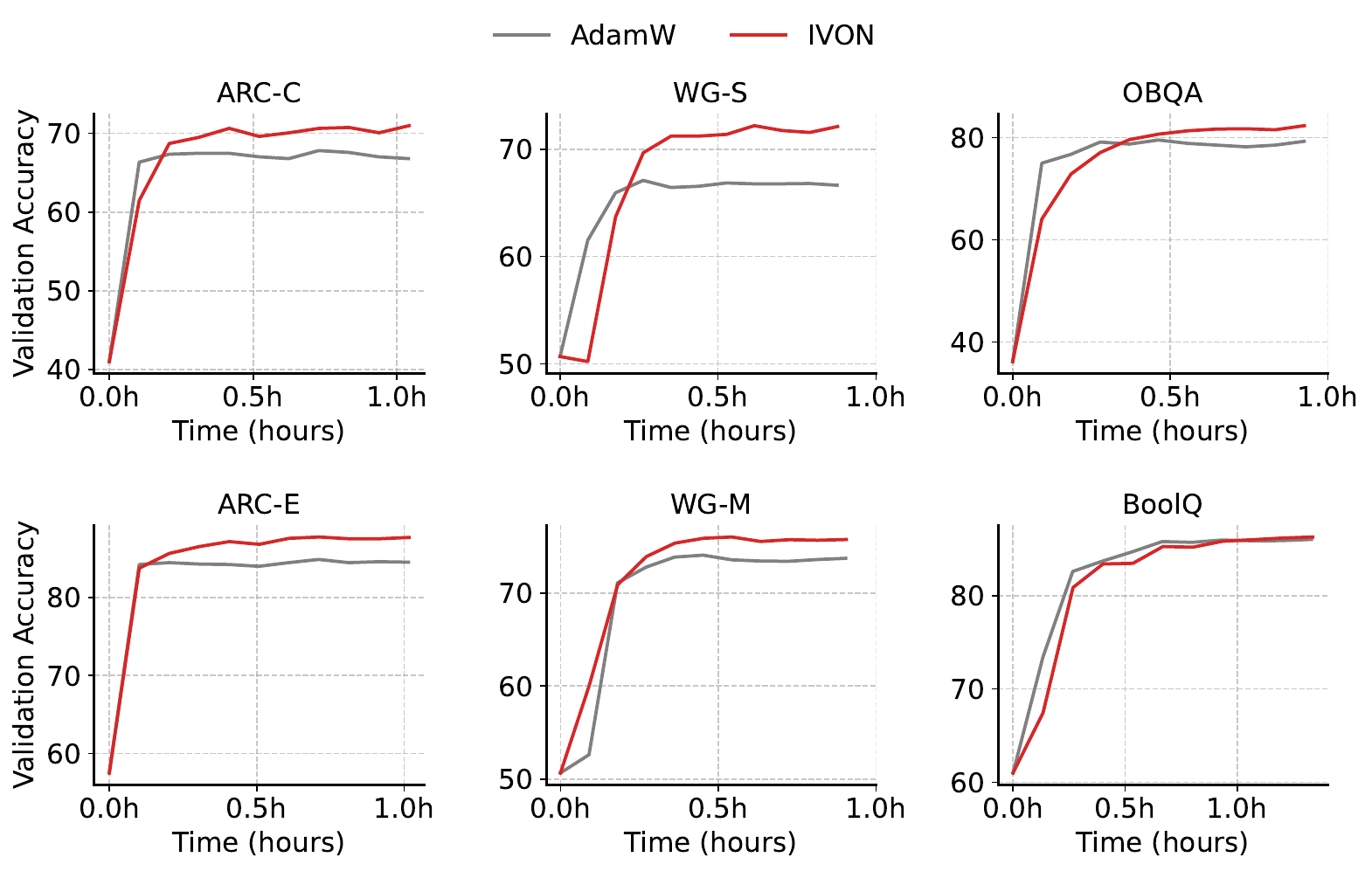}
  \caption{The training speeds of IVON and AdamW are similar. We plot validation accuracies of the two methods versus time in hours. Results are averaged over 3 runs.}
  \label{fig:curve}
\end{figure}

Finally, we observe that the per-step time overhead of IVON is negligible compared to AdamW.
We profile our training code on an NVIDIA RTX 6000 Ada GPU.
In our test run,
the forward pass, loss computation, and backward pass of a training step take in total 316.3ms on average.
As for the overhead of IVON,
the sampling procedure and the optimization step of each training step take 1.8ms and 1.0ms on average, respectively,
which is less than 1\% of the running time of a training step.
The overall training speed of IVON and AdamW are similar as shown in Figure \ref{fig:curve}.

\section{Discussion}
Our direct variational learning approach using IVON effectively improves calibration and accuracy in LoRA finetuning. Given the strong results, we hope that this work invigorates research in variational methods for LLMs. Reasons for IVON's success are not fully understood, but one hypothesis is the prevention of overfitting as the finetuning datasets are often comparably small. This 
may be attributed to the preference for simpler solutions (flatter minima) which is inherent in variational learning~\citep{HiVC93,Gr11}. 

In a broader context, several recent works consider related approaches to improve language model finetuning. Following a PAC-Bayesian framework, \citet{liu2023pac} proposes to finetune the full model using perturbed gradient descent. \citet{chen2024bayesian} uses variational learning to estimate parameter importance in adaptive LoRA (AdaLoRA)~\citep{zhang2303adaptive}. However, neither of them has been shown to work for recent billion-scale LLMs. Similar to \citet{liu2023pac}, \citet{zhelnin2024gift} shows that Gaussian noise injection can improve instruction tuning of LLMs. Different from our work, they finetune on a significantly larger instruction dataset, which is more resilient to bad calibration and overfitting. Nevertheless, these methods still demonstrate the potential of Bayesian methods and noise injection in improving LLM finetuning.

On most of the datasets,
ensemble of IVON samples outperforms IVON evaluated at the posterior mean on ECE, NLL and Brier
but at the cost of a slight decrease in accuracy.
This is perhaps due to the limited number of samples used in the ensemble.
In our experiments, we draw 10 samples for all the ensemble-based methods,
both to follow the setting in~\citet{yang2023bayesian} and to keep the computational cost manageable.
It is possible that using more IVON samples could further improve the performance of the ensemble,
which is reported in~\citet{shen2024variational} on image classification tasks.
Nevertheless, the parameter uncertainty obtained by IVON is expected to be useful for 
several downstream tasks such as sensitivity analysis~\citep{NiXu23} and model merging~\citep{DaMo23}, which will be explored in future work.

A limitation shared with other Bayesian LoRA methods~\citep{yang2023bayesian,onal2024gaussian} is that the learned posterior over the increment low-rank parameters is non-Gaussian because it is a product of two Gaussian random variables. If this is indeed a problem, a workaround could be to use a variational low-rank correction to correct the mean and variance of a Laplace approximation of the original model. \citet{van2024low} propose such a low-rank approach in the context of latent Gaussian models, and adapting these ideas to large language models may be an interesting direction for future work.

IVON also has some practical limitations. The method introduces two new hyperparameters over AdamW, namely the weighting parameter $\lambda$ and the initialization of the posterior variance $\vv$. This makes tuning IVON a bit more involved than tuning AdamW and the results depend on setting these parameters well. A good heuristic is to set $\lambda$ as small as possible while still retaining stable training and setting the posterior initialization in the order of magnitude of the final posterior
variance. The heuristic works well in practice but the method could still benefit from more principled and automatic ways to set the hyperparameters reliably. 

\begin{ack}
This work is supported by JST CREST Grant Number JPMJCR2112.
This research work has been funded by the German Federal Ministry of Education and Research and the Hessian Ministry of Higher Education, Research, Science and the Arts within their joint support of the National Research Center for Applied Cybersecurity ATHENE.
Y. Shen and D. Cremers are supported by the Munich Center for Machine Learning (MCML) and the ERC Advanced Grant SIMULACRON.
\end{ack}

\small
\bibliographystyle{plainnat}
\bibliography{references}
\normalsize


\clearpage
\section*{Supplementary Material}

\subsection*{Details on experimental setup}

Our experimental design is based on \citet{yang2023bayesian}.
We utilize the PEFT \citep{peft} library for LoRA adaptation, and apply LoRA to the query and value weights of the attention layers.
Unlike in \citet{yang2023bayesian}, we do not apply LoRA to the output layer due to numerical instability encountered in some preliminary experiments.
The base model is quantized to 8-bit precision, with LoRA weights maintained in 16-bit precision.
Finetuning is performed on a single NVIDIA RTX 6000 Ada GPU with a batch size of 4 for 10,000 steps, without gradient accumulation.

To finetune a pretrained language model which predicts the next token in a sequence for solving multiple-choice or true/false questions,
we need to wrap the text and the choice of each question with predefined prompt templates to an instruction.
We then use the pretrained model to predict the next token of the wrapped instruction,
and extract the output logits for the tokens standing for "True"/"False" or "A"/"B"/"C"/"D" choices.
For the prompt templates, we use the same ones as in \citet{yang2023bayesian}.
An example of such a prompt (used for WG-S and WG-M datasets) is as follows:
\begin{quote}
  Select one of the choices that answers the following question: \{\texttt{question}\} Choices: A. \{\texttt{option1}\}. B \{\texttt{option2}\}. Answer:
\end{quote}

\subsection*{Hyperparameters}

As for the hyperparameters of LoRA and AdamW finetuning,
we use the same settings as in \citet{yang2023bayesian},
which are also the default settings in Huggingface's Transformers \citep{wolf2020transformers} and PEFT \citep{peft} library.
For LoRA, we set the rank $r$ to 8, $\alpha$ to 16, and the dropout rate to 0.1.
For AdamW optimizer, we set the initial learning rate to $5 \times 10^{-5}$, weight decay to 0,
and use a linear learning rate scheduler which decays the learning rate to 0 at the end of the training.

Working IVON hyperparameters and guidelines for choosing them are discussed in \citet{shen2024variational}.
Still, it is not well understood how to choose them in the context of LoRA finetuning.
We empirically find that setting $\lambda$ as small as possible while still retaining stable training is a good heuristic.
To choose the initialization value $v_0$ of the posterior variance,
we track the mean value of the running average of the posterior variance for the first few training steps.
We notice that if the mean value changes significantly during the first few steps,
then the initialization value is likely too far from a reasonable one.
We follow the guideline in \citet{shen2024variational} and set the learning rate of IVON to 0.03,
Hessian momentum to $1-10^{-5}$, and clip radius to $10^{-3}$.
Finally, We summarize the hyperparameters of IVON used in our experiments in Table \ref{tab:ivon-hyperparams}.

\input{ivon-hyperparams.tex}

\end{document}

%% file: acc-calib.tex
\definecolor{verylightgray}{gray}{0.9}
\newcommand{\ptm}{\phantom{$_0$}}
\newcommand{\tab}{\hspace{0.3cm}}
\newcommand{\white}{\cellcolor{white}}
\begin{table}
    \setlength{\tabcolsep}{4pt}
    \caption{Comparison of techniques applied to finetuning/finetuned Llama-2 7B model across commonsense reasoning datasets. Results at the end of training are reported, with subscripts indicating standard error of the mean across 3 runs. We show the relative metric changes achieved by using IVON over AdamW in parentheses, with improvements in \textcolor{blue}{blue} and degradation in \textcolor{red}{red}. The methods marked with * do not require customized pipeline or additional computation during inference.}
    \label{tab:performance}
    \centering
    \resizebox{\textwidth}{!}{%
    \begin{tabular}{llrrrrrrr}
        \toprule
        \textbf{Metrics} & \textbf{Methods} & \textbf{WG-S}\ptm                 & \textbf{ARC-C}\ptm                & \textbf{ARC-E}\ptm                & \textbf{WG-M}\ptm                 & \textbf{OBQA}\ptm                 & \textbf{BoolQ}\ptm & \textbf{Average} \\ 
        \midrule
        \multirow{8}{*}{\textbf{ACC} $\uparrow$} 
        & AdamW\textsuperscript{*}          & 66.5$_{0.4}$\ptm         & 66.7$_{0.5}$\ptm  & 84.9$_{0.2}$\ptm  & 73.5$_{0.4}$\ptm  & 78.9$_{0.7}$\ptm  & 85.8$_{0.1}$\ptm  & 76.1 \\
        & \tab $+$ MC Drop                  & 66.7$_{0.4}$\ptm         & 67.3$_{0.5}$\ptm  & 84.8$_{0.4}$\ptm  & 73.7$_{0.2}$\ptm  & 79.3$_{0.5}$\ptm  & 85.9$_{0.2}$\ptm  & 76.3 \\
        & \tab $+$ LA (KFAC)                & 66.6$_{0.3}$\ptm         & 66.0$_{1.4}$\ptm  & 84.3$_{0.4}$\ptm  & 73.2$_{0.3}$\ptm  & 78.6$_{0.9}$\ptm  & 85.7$_{0.2}$\ptm  & 75.7 \\
        & \tab $+$ LA (diag)                & 66.2$_{0.3}$\ptm         & 61.2$_{1.9}$\ptm  & 81.8$_{0.5}$\ptm  & 73.3$_{0.3}$\ptm  & 79.7$_{0.8}$\ptm  & 85.7$_{0.2}$\ptm  & 74.7 \\
        & \tab $+$ SWA\textsuperscript{*}   & 69.7$_{0.6}$\ptm         & 67.2$_{1.3}$\ptm  & 85.2$_{0.1}$\ptm  & 75.6$_{0.2}$\ptm  & 79.8$_{0.5}$\ptm  & 85.5$_{0.1}$\ptm  & 77.2 \\
        & \tab $+$ SWAG                     & 69.4$_{0.6}$\ptm         & 68.4$_{1.3}$\ptm  & 85.1$_{0.2}$\ptm  & 75.2$_{0.4}$\ptm  & 80.1$_{0.1}$\ptm  & 85.2$_{0.2}$\ptm  & 77.2 \\
        \rowcolor{verylightgray}
\white  & IVON@mean\textsuperscript{*}      & {\scriptsize \textcolor{blue}{($\textcolor{blue}+5.6$)}} \bf72.1$_{0.5}$\ptm         & {\scriptsize \textcolor{blue}{($\textcolor{blue}+3.2$)}} 69.9$_{0.7}$\ptm  & {\scriptsize \textcolor{blue}{($\textcolor{blue}+2.6$)}} \bf87.5$_{0.6}$\ptm  & {\scriptsize \textcolor{blue}{($\textcolor{blue}+3.1$)}} \bf76.6$_{0.5}$\ptm  & {\scriptsize \textcolor{blue}{($\textcolor{blue}+2.0$)}} 80.9$_{0.6}$\ptm  & {\scriptsize \textcolor{blue}{($\textcolor{blue}+0.3$)}} 86.1$_{0.2}$\ptm & {\scriptsize \textcolor{blue}{($\textcolor{blue}+2.8$)}} 78.9  \\
        \rowcolor{verylightgray}
\white  & IVON                              & {\scriptsize \textcolor{blue}{($\textcolor{blue}+5.7$)}} \bf72.2$_{0.5}$\ptm         & {\scriptsize \textcolor{red}{($\textcolor{red}-0.4$)}} 66.3$_{0.6}$\ptm  & {\scriptsize \textcolor{blue}{($\textcolor{blue}+0.8$)}} 85.7$_{0.3}$\ptm  & {\scriptsize \textcolor{blue}{($\textcolor{blue}+2.9$)}} \bf76.4$_{0.6}$\ptm  & {\scriptsize \textcolor{blue}{($\textcolor{blue}+1.5$)}} 80.4$_{0.4}$\ptm  & {\scriptsize \textcolor{red}{($\textcolor{red}-0.1$)}} 85.7$_{0.2}$\ptm & {\scriptsize \textcolor{blue}{($\textcolor{blue}+1.7$)}} 77.8 \\
        \midrule
        \multirow{8}{*}{\makecell{\textbf{ECE} \\ ($\times 100$)} $\downarrow$} 
        & AdamW\textsuperscript{*}          & 32.8$_{0.5}$\ptm         & 31.4$_{0.6}$\ptm  & 14.5$_{0.3}$\ptm  & 25.3$_{0.4}$\ptm  & 19.1$_{0.8}$\ptm  & 7.6$_{0.2}$\ptm   & 21.8 \\
        & \tab $+$ MC Drop                  & 30.7$_{0.3}$\ptm         & 28.8$_{0.7}$\ptm  & 13.4$_{0.4}$\ptm  & 23.6$_{0.1}$\ptm  & 17.5$_{0.6}$\ptm  & 7.6$_{0.3}$\ptm   & 20.2 \\
        & \tab $+$ LA (KFAC)                & \bf5.2$_{2.0}$\ptm  & \bf12.4$_{2.5}$\ptm  & \bf5.4$_{1.6}$\ptm   & 11.1$_{0.2}$\ptm  & 5.5$_{0.2}$\ptm   & 3.9$_{0.1}$\ptm  & 7.3  \\
        & \tab $+$ LA (diag)                & 12.4$_{1.2}$\ptm         & \bf16.3$_{1.7}$\ptm  & 24.2$_{3.6}$\ptm  & \bf5.8$_{0.6}$\ptm   & 13.1$_{1.2}$\ptm  & 19.7$_{0.2}$\ptm  & 15.3 \\
        & \tab $+$ SWA\textsuperscript{*}   & 19.7$_{0.5}$\ptm         & 24.6$_{0.8}$\ptm  & 9.8$_{0.2}$\ptm   & 12.3$_{1.4}$\ptm  & 9.7$_{0.4}$\ptm   & 2.4$_{0.2}$\ptm   & 13.1 \\
        & \tab $+$ SWAG                     & 12.9$_{1.1}$\ptm         & \bf15.6$_{1.1}$\ptm  & \bf5.7$_{0.3}$\ptm   & 8.0$_{1.2}$\ptm   & 5.1$_{0.4}$\ptm   & \bf1.1$_{0.4}$\ptm  & 8.1 \\
        \rowcolor{verylightgray}
\white  & IVON@mean\textsuperscript{*}      & {\scriptsize \textcolor{blue}{($\textcolor{blue}-5.3$)}} 27.5$_{0.4}$\ptm         & {\scriptsize \textcolor{blue}{($\textcolor{blue}-5.6$)}} 25.8$_{0.4}$\ptm  & {\scriptsize \textcolor{blue}{($\textcolor{blue}-4.4$)}} 10.1$_{0.4}$\ptm  & {\scriptsize \textcolor{blue}{($\textcolor{blue}-2.3$)}} 23.0$_{0.5}$\ptm  & {\scriptsize \textcolor{blue}{($\textcolor{blue}-7.9$)}} 11.2$_{0.5}$\ptm  & {\scriptsize \textcolor{blue}{($\textcolor{blue}-2.0$)}} \ptm5.6$_{0.1}$\ptm & {\scriptsize \textcolor{blue}{($\textcolor{blue}-4.6$)}} 17.2  \\
        \rowcolor{verylightgray}
\white  & IVON                              & {\scriptsize \textcolor{blue}{($\textcolor{blue}-11.0$)}} 21.8$_{0.8}$\ptm         & {\scriptsize \textcolor{blue}{($\textcolor{blue}-20.7$)}} \bf10.7$_{0.4}$\ptm  & {\scriptsize \textcolor{blue}{($\textcolor{blue}-10.9$)}} \ptm\bf3.6$_{0.6}$\ptm   & {\scriptsize \textcolor{blue}{($\textcolor{blue}-3.9$)}} 21.4$_{0.5}$\ptm  & {\scriptsize \textcolor{blue}{($\textcolor{blue}-15.8$)}} \ptm\bf3.3$_{0.9}$\ptm   & {\scriptsize \textcolor{blue}{($\textcolor{blue}-5.0$)}} \ptm2.6$_{0.2}$\ptm  & {\scriptsize \textcolor{blue}{($\textcolor{blue}-11.2$)}} 10.6 \\
        \midrule
        \multirow{8}{*}{\textbf{NLL} $\downarrow$} 
        & AdamW\textsuperscript{*}          & 4.19$_{0.43}$        & 3.71$_{0.49}$         & 1.52$_{0.05}$         & 2.03$_{0.06}$         & 1.54$_{0.05}$         & 0.44$_{0.01}$ & 2.24        \\
        & \tab $+$ MC Drop                  & 3.75$_{0.33}$        & 3.25$_{0.38}$         & 1.36$_{0.07}$         & 1.85$_{0.05}$         & 1.40$_{0.04}$         & 0.43$_{0.01}$ & 2.01        \\
        & \tab $+$ LA (KFAC)                & \bf0.63$_{0.01}$        & \bf0.96$_{0.03}$         & 0.49$_{0.02}$         & 0.76$_{0.01}$         & 0.68$_{0.00}$         & 0.37$_{0.00}$ & 0.65        \\
        & \tab $+$ LA (diag)                & 0.66$_{0.01}$        & \bf1.05$_{0.05}$         & 0.70$_{0.05}$         & \bf0.57$_{0.01}$         & 0.65$_{0.01}$         & 0.47$_{0.00}$ & 0.68        \\
        & \tab $+$ SWA\textsuperscript{*}   & 0.87$_{0.02}$        & 1.34$_{0.06}$         & 0.55$_{0.00}$         & 0.63$_{0.04}$         & 0.65$_{0.02}$         & 0.34$_{0.00}$ & 0.73        \\
        & \tab $+$ SWAG                     & 0.68$_{0.02}$        & \bf1.00$_{0.04}$         & 0.46$_{0.00}$         & \bf0.56$_{0.02}$         & \bf0.55$_{0.02}$         & 0.34$_{0.00}$  & 0.60       \\
        \rowcolor{verylightgray}
\white  & IVON@mean\textsuperscript{*}      & {\scriptsize \textcolor{blue}{($\textcolor{blue}-0.69$)}} 3.50$_{0.07}$        & {\scriptsize \textcolor{blue}{($\textcolor{blue}-1.74$)}} 1.97$_{0.03}$         & {\scriptsize \textcolor{blue}{($\textcolor{blue}-0.83$)}} 0.69$_{0.00}$         & {\scriptsize \textcolor{red}{($\textcolor{red}+0.40$)}} 2.43$_{0.04}$         & {\scriptsize \textcolor{blue}{($\textcolor{blue}-0.88$)}} 0.66$_{0.02}$         & {\scriptsize \textcolor{blue}{($\textcolor{blue}-0.08$)}} 0.36$_{0.00}$  & {\scriptsize \textcolor{blue}{($\textcolor{blue}-0.64$)}} 1.60       \\
        \rowcolor{verylightgray}
\white  & IVON                              & {\scriptsize \textcolor{blue}{($\textcolor{blue}-1.94$)}} 2.25$_{0.09}$        & {\scriptsize \textcolor{blue}{($\textcolor{blue}-2.71$)}} \bf1.00$_{0.03}$         & {\scriptsize \textcolor{blue}{($\textcolor{blue}-1.12$)}} \bf0.40$_{0.00}$         & {\scriptsize \textcolor{red}{($\textcolor{red}+0.13$)}} 2.16$_{0.01}$         & {\scriptsize \textcolor{blue}{($\textcolor{blue}-1.00$)}} \bf0.54$_{0.01}$         & {\scriptsize \textcolor{blue}{($\textcolor{blue}-0.11$)}} \bf0.33$_{0.00}$ & {\scriptsize \textcolor{blue}{($\textcolor{blue}-1.13$)}} 1.11        \\
        \midrule
        \multirow{8}{*}{\textbf{Brier} $\downarrow$} 
        & AdamW\textsuperscript{*}          & 0.66$_{0.01}$        & 0.63$_{0.01}$         & 0.29$_{0.00}$         & 0.51$_{0.01}$         & 0.39$_{0.01}$         & 0.23$_{0.00}$              & 0.45         \\
        & \tab $+$ MC Drop                  & 0.63$_{0.01}$        & 0.59$_{0.01}$         & 0.28$_{0.01}$         & 0.49$_{0.00}$         & 0.37$_{0.01}$         & 0.22$_{0.00}$              & 0.43  \\
        & \tab $+$ LA (KFAC)                & \bf0.44$_{0.01}$        & 0.50$_{0.02}$         & 0.24$_{0.00}$         & 0.40$_{0.00}$         & 0.31$_{0.01}$         & \bf0.21$_{0.00}$        & 0.35         \\
        & \tab $+$ LA (diag)                & 0.47$_{0.01}$        & 0.57$_{0.03}$         & 0.35$_{0.03}$         & 0.38$_{0.01}$         & 0.33$_{0.01}$         & 0.29$_{0.00}$           & 0.40  \\
        & \tab $+$ SWA\textsuperscript{*}   & 0.48$_{0.00}$        & 0.55$_{0.01}$         & 0.25$_{0.00}$         & \bf0.37$_{0.01}$         & 0.31$_{0.01}$     & \bf0.21$_{0.00}$            & 0.36  \\
        & \tab $+$ SWAG                     & \bf0.43$_{0.01}$        & \bf0.48$_{0.01}$         & 0.23$_{0.00}$         & \bf0.36$_{0.01}$  & \bf0.28$_{0.01}$  & \bf0.21$_{0.00}$              & 0.33  \\
        \rowcolor{verylightgray}
\white  & IVON@mean\textsuperscript{*}      & {\scriptsize \textcolor{blue}{($\textcolor{blue}-0.11$)}} 0.55$_{0.01}$        & {\scriptsize \textcolor{blue}{($\textcolor{blue}-0.09$)}} 0.54$_{0.01}$         & {\scriptsize \textcolor{blue}{($\textcolor{blue}-0.07$)}} \bf0.22$_{0.01}$         & {\scriptsize \textcolor{blue}{($\textcolor{blue}-0.05$)}} 0.46$_{0.01}$         & {\scriptsize \textcolor{blue}{($\textcolor{blue}-0.09$)}} 0.30$_{0.01}$         & {\scriptsize \textcolor{blue}{($\textcolor{blue}-0.02$)}} \bf0.21$_{0.00}$    & {\scriptsize \textcolor{blue}{($\textcolor{blue}-0.05$)}} 0.40          \\
        \rowcolor{verylightgray}
\white  & IVON                              & {\scriptsize \textcolor{blue}{($\textcolor{blue}-0.18$)}} 0.48$_{0.01}$        & {\scriptsize \textcolor{blue}{($\textcolor{blue}-0.16$)}} \bf0.47$_{0.01}$         & {\scriptsize \textcolor{blue}{($\textcolor{blue}-0.08$)}} \bf0.21$_{0.01}$         & {\scriptsize \textcolor{blue}{($\textcolor{blue}-0.07$)}} 0.44$_{0.01}$         & {\scriptsize \textcolor{blue}{($\textcolor{blue}-0.11$)}} \bf0.28$_{0.01}$         & {\scriptsize \textcolor{blue}{($\textcolor{blue}-0.02$)}} \bf0.21$_{0.00}$      & {\scriptsize \textcolor{blue}{($\textcolor{blue}-0.10$)}} 0.35         \\
        \bottomrule
    \end{tabular}
    }
\end{table}

%% file: ivon-hyperparams.tex
\begin{table}
    \caption{IVON hyperparameters used in experiments.}
    \label{tab:ivon-hyperparams}
    \centering
    \begin{tabular}{l S[table-format=1e1] S[table-format=1e1] S[table-format=1e1] S[table-format=1e1] S[table-format=1e1] S[table-format=1e1]}
        \toprule
        \textbf{Hyperparameter} & \textbf{WG-S} & \textbf{ARC-C} & \textbf{ARC-E} & \textbf{WG-M} & \textbf{OBQA} & \textbf{BoolQ} \\ 
        \midrule
        Effective sample size & 1e7 & 1e6 & 1e6 & 1e8 & 1e6 & 1e7 \\
        Hessian initialization & 3e-4 & 1e-3 & 1e-3 & 3e-4 & 1e-3 & 3e-4 \\
        Learning rate & \multicolumn{6}{c}{$0.03$} \\
        Gradient momentum & \multicolumn{6}{c}{$0.9$} \\
        Hessian momentum & \multicolumn{6}{c}{$1-10^{-5}$} \\
        Clip radius & \multicolumn{6}{c}{$10^{-3}$} \\
        \bottomrule
    \end{tabular}
\end{table}